Shades of meaning:

Uncovering the geometry of ambiguous word representations through contextualised language models

Benedetta Cevoli,[1] Chris Watkins,[2] Yang Gao[2] & Kathleen Rastle[1]

1. Department of Psychology, Royal Holloway, University of London
2. Department of Computer Science, Royal Holloway, University of London

RUNNING HEAD: SHADES OF MEANING

Correspondence to:

Kathy.Rastle@rhul.ac.uk

## Abstract

Lexical ambiguity presents a profound and enduring challenge to the language sciences. Researchers for decades have grappled with the problem of how language users learn, represent and process words with more than one meaning. Our work offers new insight into psychological understanding of lexical ambiguity through a series of simulations that capitalise on recent advances in contextual language models. These models have no grounded understanding of the meanings of words at all; they simply learn to predict words based on the surrounding context provided by other words. Yet, our analyses show that their representations capture fine-grained meaningful distinctions between unambiguous, homonymous, and polysemous words that align with lexicographic classifications and psychological theorising. These findings provide quantitative support for modern psychological conceptualisations of lexical ambiguity and raise new challenges for understanding of the way that contextual information shapes the meanings of words across different timescales.

Most words in English (as in other languages) have multiple meanings. The word *bank* is a prototypical example: it can refer to a financial institution, the edge of a river, or a manoeuvre in an aeroplane. Words like *bank* that have multiple unrelated meanings are known as homonyms. It is also possible for words to have multiple related senses as in the word *line* (e.g., behind the enemy line, a new line of attack, get the ball over the line, draw a line); such words are known as polysemes. Debate for decades has asked why languages admit ambiguity (Wittgenstein, 1958), and what the consequences of it are for language processing (Johnson-Laird, 1987; Swinney, 1979). Modern linguistic perspectives argue that the reuse of wordforms to represent different meanings in context may eliminate redundancy (Piantadosi et al., 2012) and enhance communicative efficiency (Gibson et al., 2019). However, psychological perspectives have generally considered lexical ambiguity as posing a difficult problem for learning (Fang et al., 2017) and as an obstacle to successful comprehension (Rodd, 2020).

The present study is focused on the nature of ambiguous words' representations, and specifically, how the geometry of these representations reflects the text contexts in which ambiguous words occur. This is an important undertaking because leading psychological theories of ambiguity have *assumed* that homonyms and polysemes have a particular geometry without providing evidence of how this might arise (Rodd, 2020). We study the representations of ambiguous words in the language model BERT (Devlin et al., 2019). This model represents word meanings as high-dimensional vectors derived from analysis of the way that words occur in very large text corpora. BERT offers a substantial advance on earlier language models such as LSA (Landauer & Dumais, 1997) and word2vec (Mikolov et al., 2013) because instead of combining all possible meanings of a word into a single vector representation, it represents words in a context-dependent manner. That is, each word in this model is represented by high-dimensional vectors that change dynamically based on the surrounding context. This property makes the model ideally suited to studying the representations of words that have multiple meanings.

## Psychological Perspectives on Lexical Ambiguity

The past 40 years of psychological research on lexical ambiguity has been dominated by the question of how the appropriate meaning of an ambiguous word is selected sufficiently rapidly to permit fluent comprehension. The earliest "exhaustive access" models proposed that all meanings of a word are activated in sentence processing no matter the context (e.g. Swinney, 1979) and therefore must be disambiguated online. These theories were replaced by "selective access" models proposing that sufficiently constraining contexts can prevent the costly activation of inappropriate meanings (Tabossi, 1988). The field ultimately settled on an intermediate theory in which a range of factors including context can influence activation of the different meanings of an ambiguous word (Duffy et al., 1988). The important point for our purposes is that these theories all assumed a binary characterisation of ambiguity – a word is ambiguous or it's not – and that words have a discrete number of meanings or senses. These studies typically quantified the ambiguity status of a word by asking participants to rate the number of meanings associated with it (Hino et al., 2002; Pexman et al., 2004) or by counting the number of dictionary definitions assigned to it (Klein & Murphy, 2001; Rodd et al., 2002).

Modern research has moved to a more nuanced characterisation of ambiguity, in which word meanings are assumed to vary continuously as a function of the contexts in which they occur. In this framing, the different meanings or senses assigned to ambiguous words in dictionary entries reflect an artificial structure imposed by lexicographers on this continuous variation (Hoffman et al., 2013). This more nuanced perspective resonates with modern theories of ambiguity resolution typically conceived within a distributed-connectionist framework (e.g. Armstrong & Plaut, 2016; Kawamoto et al., 1994; Rodd et al., 2004). In the most recent of these theories (Rodd, 2020), the meanings of words are characterised as distributed representations, in which familiar meanings occupy a single point in a high-dimensional semantic space. In the case of unambiguous words such as 'frog', the mapping between form and meaning is straightforward, with a single form mapping onto a single high-dimensional semantic representation. In contrast, in the case of ambiguous words such as homonymous 'bark' or polysemous 'shade', a single form maps onto two or more high-dimensional semantic representations (see Figure 1). This theory proposes that resolving the meaning of an ambiguous word reflects a dynamic settling process between high-dimensional semantic representations of the different meanings of that word, with processing penalties arising as a result of settling into that stable state (Rodd, 2020) .

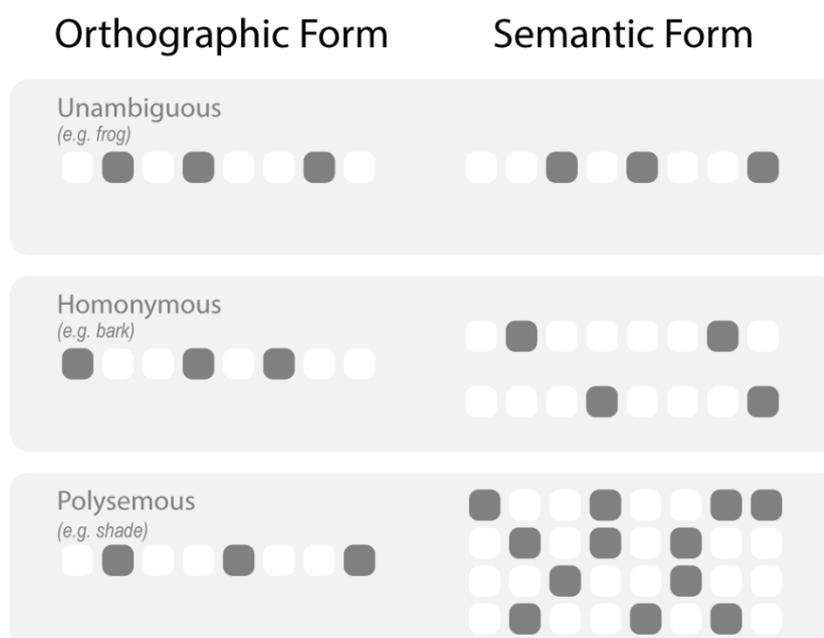

**Figure 1.** Illustration of orthographic and semantic representations of homonymous and polysemous words (redrawn from Rodd, 2020).

This representational structure captures the linguistic distinction between homonymy and polysemy yet reflects that these forms of ambiguity are on a continuum (Rodd, 2020). Words are homonymous to the extent that they map onto multiple non-overlapping semantic representations, representing more distant points in the semantic space. Words are polysemous to the extent that they map onto multiple overlapping semantic representations, representing nearer points in the semantic space. This representational structure captures variation in the ease with which the different meanings and senses of ambiguous words can be separated. It is trivial to separate the different meanings of a homonym like 'bark'; and indeed, these representations are assumed to have low overlap. However, researchers for decades have recognised the difficulties associated with segmenting polysemous words like 'run' into distinct senses, sometimes taking teams of lexicographers many months to undertake what is ultimately a subjective task (see e.g. Gilliver, 2016; Kilgarriff, 1997; McDonald & Shillcock, 2001; Tuggy, 1993 for discussion). The difficulty in distinguishing these fine gradations is reflected in representations that have very high overlap. However, the classification of ambiguous words as homonymous or polysemous is a matter of degree; there is no arbitrary point at which the distance in semantic space would trigger one or the other classification (Rodd, 2020). Likewise, while "hybrids" (Armstrong & Plaut, 2016) such as *port* that have features of both homonyms and polysemes might pose difficulty in a binary classification, they are represented straightforwardly on this scheme.

The representational structure illustrated in Figure 1 therefore captures important information about the extent to which the different meanings and senses of words are related. This structure is also central to understanding aspects of the online processing of ambiguous words. For example, Rodd et al. (2002) showed that homonyms are recognised more slowly than unambiguous words, while polysemes are recognised more quickly than unambiguous words. In a distributed-connectionist network implementing the representational structure illustrated in Figure 1, Rodd et al. (2004) showed that the homonymy disadvantage arises because of the processing cost associated with moving the network from an initial blend state comprising multiple unrelated meanings to a single stable representation. In contrast, the polysemy advantage arises because settling behaviour in the network benefits from broad, deep semantic attractors arising from large clusters of overlapping semantic representations. Thus, the move to thinking about ambiguity in terms of continuous variation in meaning (as opposed to a discrete, binary phenomenon) offered not only a richer understanding of the meaningful structure of ambiguous words but also a tighter link to processing than had been possible previously.

Despite the advances that this representational scheme provides for our understanding of ambiguity, one important gap is that it is unclear how this structure arises. The illustrations provided in Figure 1 are just schematics expressing an abstract hypothesis. Likewise, relevant distributed-connectionist simulations have simply built the proposed differences between unambiguous, homonymous, and polysemous words into their representations (Rodd et al., 2004; see also Armstrong & Plaut, 2008). It seems clear that this structure must arise through recent and long-term experience with ambiguous words (Rodd, 2020), but greater specificity would be needed to advance the type of computational account that has been so fruitful in cognitive science.

Lexical Ambiguity in Language Models

Our work uses distributional semantic modelling to investigate to what extent this representational structure may emerge from simple statistics pertaining to the usage of ambiguous words in text. Distributional models derive high-dimensional semantic spaces through the analysis of word co-occurrence in very large text corpora (e.g. Burgess, 1998; Landauer & Dumais, 1997; Mikolov et al., 2013). Representations of individual words are points in this high-dimensional space, and the distance between points reflects the degree to which those words appear in similar contexts (and hence, have similar meanings). Hoffman et al. (2013) hypothesised that ambiguous words may occur in more diverse contexts than unambiguous words: for example, homonym 'bank' may occur in very different contexts pertaining to rivers and money while unambiguous 'perjury' will likely be restricted to legal scenarios. If so, then it is possible that the high-dimensional semantic representations of these words will reflect those different clusters of contextual usage.

Hoffman et al. (2013) proposed a method for using distributional semantic modelling to measure the diversity of a word's contexts. They used LSA (Landauer & Dumais, 1997) to derive high-dimensional vectors for each context in which a word occurs (using 1000-word contexts) and then measured the pairwise cosine similarity between the context vectors. Higher pairwise cosine similarity indicates less diverse (i.e. more similar) contexts. It is important to note that the high-dimensional vectors being used here are not of each *word*; instead, they are for the whole 1000-word context (and thus every word in that 1000-word context is represented by the same high-dimensional vector). Hoffman et al. (2013) did not report whether ambiguous words had more diverse contexts on this measure than unambiguous words. However, Cevoli et al. (2021) used their approach to derive diversity measures for stimuli used in two studies investigating adults' processing of homonymous, polysemous, and unambiguous words (Armstrong & Plaut, 2016; Rodd et al., 2002). Cevoli et al. (2021) found no evidence that ambiguous words (homonyms and polysemes) occur in more diverse contexts than unambiguous words on this LSA-based measure.

In response, Hoffman et al. (2021) argued that Cevoli et al. (2021) had weighted their context vectors by the singular values, and if this processing step is not taken then polysemous words in the two studies simulated (Armstrong & Plaut, 2016; Rodd et al., 2002) do show greater diversity in their contexts than unambiguous words. However, there was still no evidence from their LSA-based measure that homonyms occur in more diverse contexts than unambiguous words. Moreover, using unweighted vector representations (meaning that each dimension of the vector is assumed to contribute equally to the meaning) is non-standard in LSA; for example, "*Therefore, to do a comparison between any two documents, or a document and a pseudo-document, the document vectors, or pseudo-document vector, scaled by the singular values are used to compute the similarity measure.*" (Martin & Berry, 2013, p. 51). Vectors are weighted in LSA to reflect the most important dimensions; failure to do so leaves representations highly vulnerable to fluctuations in dimensionality. This means that if a small proportion of dimensions in a representation are particularly important, then the influence of those dimensions will be much weaker in an unweighted 1000-dimensional vector as compared to an unweighted 100-dimensional vector. The use of weighted vectors (as in Cevoli et al., 2021) avoids this problem.

Beekhuizen et al. (2021) took a different approach to probing whether distributional semantic representations derived from analysis of large text corpora can capture the ambiguity structure of a word. Instead of investigating the nature of the *contexts* in which ambiguous and unambiguous words occur, they attempted to investigate the nature of word representations themselves. They approached this by measuring the similarity between word2vec (Mikolov et al., 2013) representations of homonymous, polysemous, and unambiguous targets and a series of *probes* (e.g. words taken from dictionary definitions of targets). They hypothesised that the similarity between unambiguous targets and probes should be greater than the similarity between ambiguous targets and probes; and further, that the similarity between polysemous targets and probes should be greater than the similarity between homonymous targets and probes. These hypotheses are illustrated in Figure 2 which depicts the similarity structure between three targets and their probes (Beekhuizen et al., 2021). Results were numerically in line with Beekhuizen et al.'s prediction although the pattern of statistical significance varied across different types of probes used. This analysis begins to suggest that the ambiguity structure of words may be recovered through analysis of relatively superficial properties of text, and hence may be learned through experience with text. However, while this approach did not focus on contexts, the inferences made in this research were based (at least partly) on the semantic representations of probes, rather than representations of the target words. More persuasive evidence would arise from analysis of the meaning structure of target words themselves.

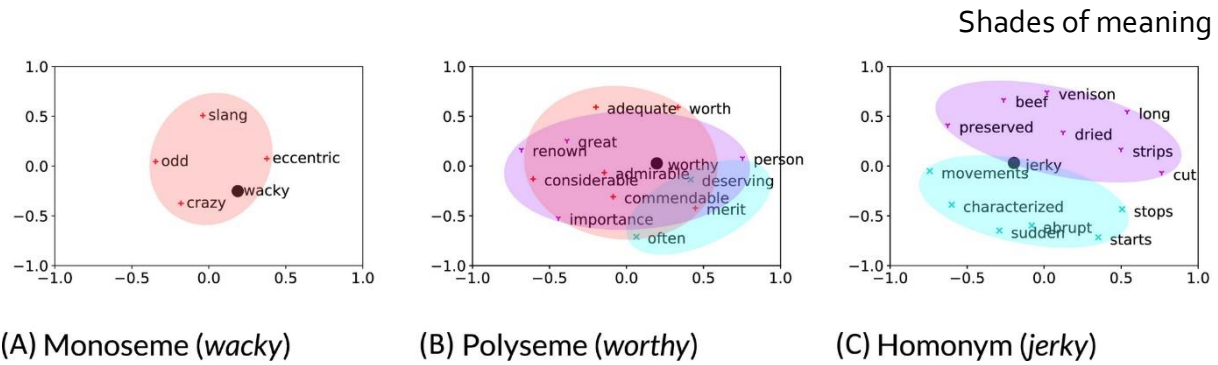

(A) Monoseme (*wacky*)    (B) Polyseme (*worthy*)    (C) Homonym (*jerky*)

**Figure 2**. Multidimensional scaling plots showing similarity structure of three targets and their probes. The figure is reproduced from Beekhuizen et al. (2021). (We will gain permission to reproduce the figure from the publisher if this manuscript is accepted).

These initial investigations of ambiguity in distributional semantic models have used 'static' models such as LSA (Landauer & Dumais, 1997) and word2vec (Mikolov et al., 2013). These models are static in the sense that they derive a single high-dimensional semantic representation for each word, combining all possible meanings of a word into that single representation. Though some researchers have been optimistic that these representations may capture multiple meanings (Arora et al., 2018; Beekhuizen et al., 2021; Mu et al., 2017), lexical ambiguity has long been recognised to pose a challenge for these models. For example, a recent review of distributional semantic models noted

> "*The prevailing objective of representing each word type as a single point in the semantic space has a major limitation: it ignores the fact that words can have multiple meanings and conflates all these meanings into a single representation.*" (Camacho-Collados & Pilehvar, 2018, *p. 5*)

More recent 'dynamic' models such as BERT (Devlin et al., 2019) offer new opportunities to discover how a word's ambiguity structure can be derived from the usage of words in text. BERT is trained on vast text corpora on two unsupervised prediction tasks: a task in which the model learns to predict the next word and a task in which the model learns to predict the likelihood of an upcoming sentence. The key advance for our purposes is that BERT learns context-dependent representations of individual words; hereafter, we refer to these as *contextualized embeddings*. That is, each word is described by high dimensional vectors that change dynamically based on the surrounding context. This property allows BERT to represent *specific instances* of a word within high-dimensional semantic space. Thus, it is at least possible that the different meanings of ambiguous words might occupy different regions of that space, in much the same way as proposed in modern psychological theories of ambiguity (Rodd, 2020).

BERT and similar dynamic language models have received a great deal of attention in respect of natural language applications but are only just beginning to be inform psychological questions (e.g. Cevoli et al., 2022). However, research using a recurrent neural network (RNN) based on a simpler predecessor to BERT (LSTM; Merity et al., 2017) provided initial evidence that the different senses of ambiguous words may be recovered from properties of the way that these words are used in context (Li & Joanisse, 2021). This work showed that context-dependent representations in the model capture pre-defined sense labels of ambiguous words significantly better than a random distribution of sense labels. This result indicates that the model has learned *something* about these different meanings through the analysis of low-level properties of text usage. However, showing that the representations encode the right meaning significantly better than a random distribution of meanings is a weak test of the precision of these representations. Moreover, all but one of the words considered had only two or three senses (unlike ambiguous words such as *line* or *shade* with multiple related senses). Nevertheless, this work provides a promising foundation for a deeper investigation of the extent to which ambiguity structure emerges in representations acquired in more powerful natural language models that analyse usage patterns in very large and diverse text corpora.

The initial work in this domain is important because it suggests that the types of representations envisaged in psychological theories of ambiguity (Rodd, 2020) may be derived through analysis of relatively simple text usage statistics. Thus far, there has been little discussion as to how those representations might be acquired, but clearly, understanding what might give rise to representations that convey a word's ambiguity structure is vital in assessing those theories. It may be that humans learn these representations in a different way from the models or use a broader range of information than is available to the models. Indeed, there is already a substantial literature describing the limitations of these models in capturing human language understanding (e.g. Bender et al., 2021; Kwon et al., 2019; Marcus & Davis, 2019; McClelland et al., 2019). However, if we were to show that the ambiguity structure of a word could be recovered from unsupervised learning of relatively low-level information in text, then that would indicate that psychological theories of ambiguity are not *required* to postulate more complex mechanisms for the acquisition of these representations.

### Data Availability

Materials, data, and analysis scripts for this are available in the Open Science Framework (https://osf.io/435jw).

### Simulation 1

The aim of this simulation was to examine whether BERT representations show evidence of a word's ambiguity structure. Our approach was to investigate the semantic spread of language model representations of words that have previously been classified as homonymous, polysemous, or unambiguous in published word recognition studies (Armstrong & Plaut, 2016; Rodd et al., 2002). We sought in the first instance to determine whether there is systematic variation in the high dimensional representations of these different types of words in context. Our hypothesis was that representations of homonymous and polysemous words should show greater semantic spread than unambiguous words. That is, the representations associated with homonymous and polysemous words should be further apart (reflecting their multiple meanings) than representations of unambiguous words.

We acquired multidimensional representations of homonymous, polysemous, and unambiguous words in different contexts (*contextualised embeddings*), using the BERT natural language model (Devlin et al., 2019). BERT is a multi-layer transformer network that calculates multidimensional representations of each token in an input sequence. It is important to recognise that we are investigating the nature of these high-dimensional representations themselves. We are not deriving representations of whole contexts (as in Hoffman et al., 2013, 2021; Cevoli et al., 2021), and we are not mapping the similarity structure of target words against probes (Beekhuizen et al., 2021). Instead, we are investigating the *structure of word embeddings in context* (i.e. multidimensional representations of specific words in different contexts). Such investigations have only recently become possible through advances in natural language models (e.g. Devlin et al., 2019).

## Method

We focused on two studies of human word recognition that reported effects of lexical ambiguity for which stimuli were available (Armstrong & Plaut, 2016; Rodd et al., 2002). Working with these two studies also allowed an informal comparison with the results of Cevoli et al. (2021), who found no difference in the diversity of contexts in which homonymous, polysemous, and unambiguous words occur using the LSA-based measure of Hoffman et al. (2013). Rodd et al. (2002) included a total of 182 items in their Experiment 1 (124 homonymous and 58 unambiguous words) and 256 in their Experiment 2 (128 with few vs. many senses and 128 with one vs. many meanings). Likewise, 374 items were used in Armstrong and Plaut (2016) of which 91 were homonymous, 96 polysemous, and 93 unambiguous (a further 94 hybrid words were not considered in our study). In total, we selected a total of 641 items from these studies (259 homonymous, 140 polysemous, and 242 unambiguous).

We derived contexts in which these target words occur using the British National Corpus (BNC; The British National Corpus, 2007). Contexts were defined as the closest number of sentences to a 100-word window. We focused on sentences as the building blocks of contexts in order to preserve the linguistic structure of natural language. Starting from the sentence in which the target word occurs, we compared the length of the target sentence (current context) and the length of its immediate broader context extended to the sentence before and after (broader context), a total of three sentences in the first instance. We then determined whether the current context or the broader context was closest to our desired 100-word window. If the broader context was closest, then we repeated the process (adding an additional sentence before and after the current context) until the current context was closest to 100 words. This process resulted in a set of naturalistic contexts with varying lengths of approximately 100-words (M=99.94; SD=16.34). In total, we retrieved 2,589,921 contexts from the BNC (an average of 3,368 contexts per target word; see supplementary Table 1 in the OSF site for this project for distribution of contexts across conditions).

We obtained embeddings of each target word *in each of these contexts* using a pre-trained version of BERT (model version: *bert-large-uncased*; PyTorch Transformers library; Wolf et al., 2019; Devlin et al., 2019). To derive contextual embeddings for our target words, we submitted all 2,589,921 contexts (as defined above) as input to the pre-trained model. We then extracted multidimensional representations of each target word in each context (i.e. the contextual embeddings) from the final layer of the language model. Each of the resulting 2,589,921 contextual embeddings had 1,024 dimensions.

We ran simulation analyses of lexical ambiguity studies (Rodd et al., 2002; Armstrong & Plaut, 2016) to determine whether the ambiguity structure of target words in these studies was apparent in the sample of BERT contextual embeddings obtained for each of these target words. For each target word, all its contextual embeddings were used to derive a measure of a word's semantic spread (hereafter, *embedding diversity*). This measure was computed by taking the pairwise cosine distance between each of the contextual embeddings of a word and every other contextual embedding of that same word, and then averaging these pairwise distance values (methods adapted from Hoffman et al., 2013; see also Hsiao & Nation, 2018; Cevoli et al., 2021). These simulation analyses provide a first step to understanding the extent to which the different meanings of homonymous and polysemous words can be recovered via unsupervised learning of statistical relationships in text.

## Results

Rodd et al. (2002) investigated the effect of lexical ambiguity on visual lexical decision performance in a multi-experiment study using a regression and factorial design (their Experiments 1 and 2, respectively; see also Cevoli et al., 2021 for replication using mega-study data). Simulation analyses using items from Rodd et al. (2002) revealed a significant effect of the number of senses on embedding diversity (Exp 1: $\beta = 0.46, SE = 0.01, t = 6.5, p < 0.001$; Exp 2: $F(1,122) = 79.41, p < 0.001$, $\eta_p^2 = 0.39$). We also observed a significant effect of the number of meanings on embedding diversity (Exp 1: $\beta = 0.27, SE = 0.02, t = 4.00, p < 0.01$; Exp 2: $F(1,122) = 4.75, p < 0.05$, $\eta_p^2 = 0.04$). Similar results were obtained when we quantified embedding diversity for the items used by Armstrong and Plaut (2016). Unambiguous words had significantly lower embedding diversity than both homonymous words ($\beta = 0.27, SE = 0.01, t = 3.94, p < 0.001$) and polysemous words ($\beta = 0.46, SE = 0.02, t = 7.05, p < 0.001$; see Figure 3). These results suggest that the contextual embeddings of words with a larger number of senses or meanings tend to show greater spread than those with a lower number of senses or meanings.

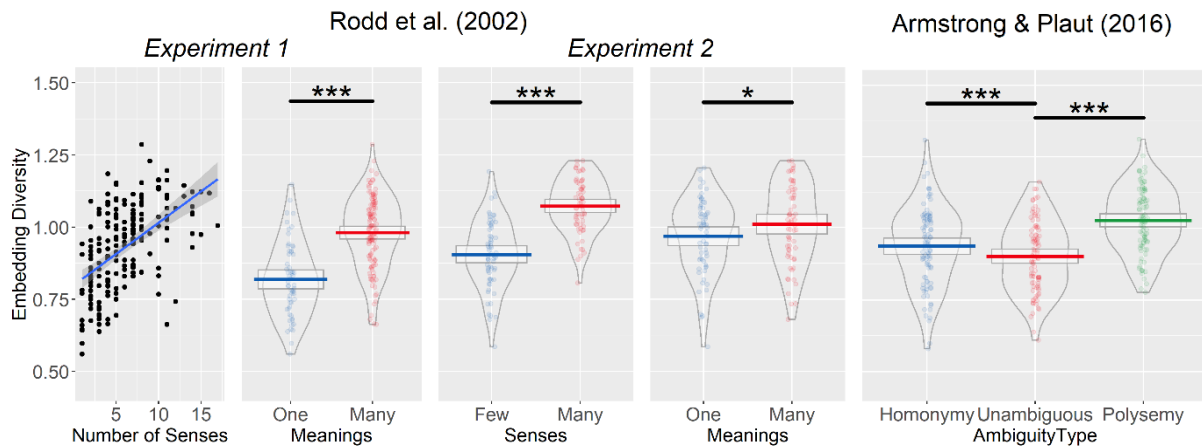

**Figure 3.** Results of the simulation analyses of Rodd et al. (2002) and Armstrong and Plaut (2016) on embedding diversity values. Words with a higher number of senses or meanings (based on dictionary entries) have greater semantic spread.

## Discussion

This work provides initial evidence that the ambiguity structure of a word may be captured in the multidimensional lexical representations learned through the analysis of statistical relationships in text. Using a natural language model sensitive to contextual variation, we have shown that multidimensional representations of homonymous and polysemous words show greater semantic spread than those for unambiguous words. We believe that this is the first demonstration that high-dimensional representations of homonyms and polysemes learned through analysis of text statistics show a different geometry to high-dimensional representations for unambiguous words.

The data described in Figure 3 suggest that semantic spread for polysemous words may be even greater than that for homonymous words. This result seems inconsistent with the predictions described in Figures 1 and 2, in which polysemous words are characterised by multiple overlapping representations. However, previous work has postulated that polysemous words differ greatly in the extent to which their different senses are related to one another (e.g. Bartsch, 1984; Klepousniotou et al., 2008), a possibility that makes room for item-level effects. Likewise, although the original authors of the target studies classified their words as homonymous and polysemous based on the structure of dictionary entries, ultimately the decision to treat a semantic concept as a new sense as opposed to a new meaning in the dictionary is a subjective one. Thus, we do not believe that strong inferences should be drawn regarding the apparently greater semantic spread of polysemous words than homonymous words in these studies. Further work is required to determine whether this pattern holds across other stimulus sets (and if so, why).

The fact that contextual embeddings of homonymous and polysemous words showed greater semantic spread than those for unambiguous words suggests that (a) the contexts in which homonymous and polysemous words occur are likely to be more diverse than those in which unambiguous words occur; and (b) that this variation may be recovered through the relatively low-level input available to modern natural language models (Devlin et al., 2019). However, while the present results show that the contextual embeddings of homonymous and polysemous words show greater semantic spread than those for unambiguous words, they do not demonstrate that this spread is meaningful (i.e. that contextual embeddings of a word that are further apart represent different meanings of that word). Moreover, they do not provide much information about the geometry of a word's representations; for example, whether contextual embeddings of homonyms differ from those of polysemous words in the way postulated by Rodd (2020; see Figure 1). Consequently, we next took a deep dive into the representational structure of homonymous and polysemous words, using a case study approach. This type of detailed investigation of language model representations is rarely undertaken but critical for understanding why particular effects arise.

### Simulation 2

Our analyses thus far suggest that the contextual embeddings of homonymous and polysemous words show significantly greater semantic spread than those of unambiguous words; however, this result says nothing about whether this spread is meaningful. We therefore went 'under the bonnet' to analyse the distribution of contextual embeddings for the homonymous word *bark* and the polysemous word *shade* (both taken from Rodd et al., 2002). Our first aim in this analysis was to assess the extent to which multidimensional representations learned through text input capture the multiple different meanings of words in context. Our second aim was to investigate whether the geometry of these multidimensional representations align with characterisations of homonymous and polysemous words described in previous research (e.g. Beekhuizen et al., 2021; Rodd, 2020; see Figure 1).

## Method

We conducted an in-depth analysis of the contextual embeddings of *bark* and *shade* derived from the same pre-trained version of BERT used in the previous section (model version: *bert-large-uncased*; PyTorch Transformers library; Wolf et al., 2019; Devlin et al., 2019).

One immediate challenge in trying to analyse the nature of meaningful information in these contextual embeddings is that we require ground truth data regarding the meanings of *bark* and *shade* in each of the contexts under scrutiny. Obtaining such data is relatively straightforward (although laborious) for homonymous words; for example, one could ask participants to decide if the word *bark* in a particular context referred to the cry of an animal or the protective sheath of a tree. Estimating ground truth meaning for polysemous words is far less straightforward because these words may have hundreds of interrelated meanings. One might be tempted to use a dictionary as an objective source of ground-truth meaning, but different dictionaries disagree on where meaning boundaries lie, perhaps according to whether the chief lexicographer is a 'lumper' or 'splitter' (Allen, 1999). Allen (1999) writes "…*there are no absolutes in dictionary writing*" (p. 62) and gives an example of three dictionary treatments of the verb *make* yielding 7, 24 and 35 meanings. In light of the difficulty that polysemous words pose to lexicographers, it seems unlikely that robust ground truth data could be obtained by asking samples of undergraduates or adults from the general public to articulate these distinctions.

We developed an innovative solution to this problem capitalising on the fact that the different meanings of homonymous and polysemous words in English are frequently represented by distinct words in other languages. In the case of *bark*, the Italian translation *abbaiare* typically refers to a dog bark, while *corteccia* is used to refer to tree bark. Similarly, *shade* is translated into 16 different Italian words according to Google Translate (Google, 2021). Our approach was to obtain Italian translations for contexts including *bark* and *shade*, and to use the translations of these words in each context as an estimate of their ground truth meaning. We chose Italian because research at the point of our study indicated that the quality of translations from English to Italian is the highest of any language that Google Translate supports (Aiken, 2019). We recognize that this approach may not produce the precise distinctions visible to a lexicographer; and we also recognize that translation to a different language may produce slightly different meaning categories (perhaps akin to the use of different dictionaries). Our aim was to source *estimates* of ground truth meaning in an efficient manner that could be applied rapidly to thousands of contexts in future.

We randomly selected contexts from the BNC (The British National Corpus, 2007) in which *bark* and *shade* occurred following the same procedure described in the first set of simulations (N=500 contexts per word). Automatic Italian translations were obtained for these contexts from Google Translate using the python Deep Translator library (Baccouri, 2020). This procedure resulted in three unique Italian translations for *bark* and 16 for *shade*. The translation of each target word in context was used as a proxy for its ground truth meaning; for example, the translation of *bark* to *abbaio* was taken to reflect the 'dog bark' meaning, while translation to *corteccia* was taken to reflect the 'tree bark' meaning.

We validated these estimates using meaning judgments from three British English speakers who were naïve to the purposes of the study. These participants were asked to select the correct meaning of the target words *bark* and *shade* in each context, using meaning choices based on the labels identified through the translation process back-translated to English (these labels are available in Supplementary Table 2 on the OSF site for this project). Instances in which two different Italian translations mapped onto indistinguishable meanings in English were combined into one label. This was the case for one pair of Italian translations for *bark* (*abbaio* and *latrato* both indicate a "dog bark") and two pair of Italian translations for *shade* (*un po'* and *leggermente* both refer to the meaning 'a little'; and *proteggere* and *riparare* both refer to the meaning 'to protect from sun'). This process resulted in two translation labels for *bark* and 14 for *shade*; human raters were also given the option to select 'other' in both cases. Results showed satisfactory inter-rater reliability between the automatic translations and each of the human raters (*bark*: average $\alpha = 0.93$ with 2 meaning choices plus 'other'; *shade*: average $\alpha = 0.85$ with 14 meaning choices plus 'other'). We were therefore satisfied that the automatic translations provided reasonable ground truth data from which to evaluate the information encoded in BERT's contextual embeddings of *bark* and *shade*.

We ran the same 1000 contexts through BERT and extracted the contextual embeddings for *bark* and *shade*, using the procedure described for the first simulation. Our analyses sought to determine whether the contextual embeddings of *bark* and *shade* showed evidence of their distinct ground truth meanings.

<div align="center">

**Results**

</div>

We visualised the multidimensional vectors for the words *bark* and *shade* with t-Distributed Stochastic Neighbour embedding (t-SNE; Van Der Maaten & Hinton, 2008). Figure 4 shows the t-SNE plot of BERT's contextual embeddings of the words *bark* and *shade*; similar usages of words are typically illustrated by points close to each other in the t-SNE space while different uses typically have points further apart. Points in the t-SNE space for *bark* and *shade* have been coloured according to their Italian translations (i.e. our ground truth labels).

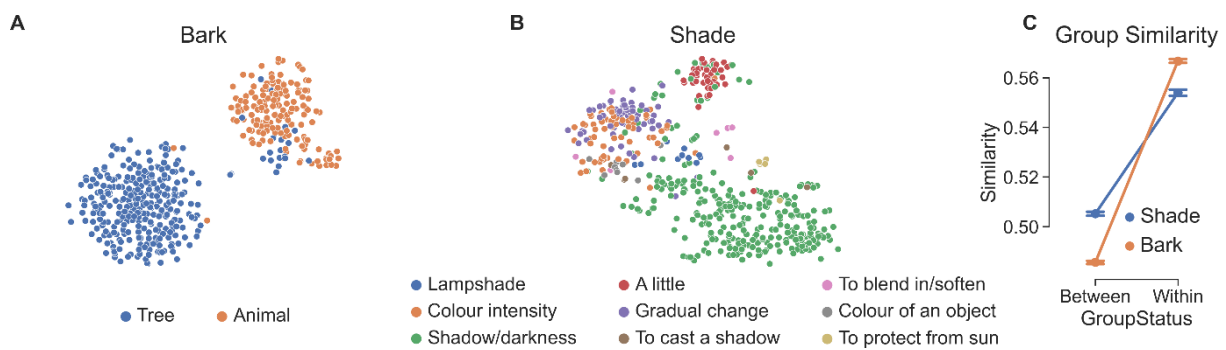

**Figure 4.** t-SNE plot of contextual embeddings of the words *bark* (A) and *shade* (B) coloured by their translation labels. Each point refers to a different contextual embedding (i.e. multidimensional representation from a different context). Points closer to each other in the t-SNE space represent contextual embeddings that are more similar to one another. For illustration purposes, only the 9 most frequent Italian translations of *shade* were selected. (C) Intra- and inter-group cosine similarity of the contextual embeddings of *bark* and *shade*. Error bars indicate 99% confidence intervals.

It is immediately apparent that the contextual embeddings of *bark* and *shade* extracted from BERT contain important disambiguating information. The multidimensional representations show substantial separation between the 'dog bark' and 'tree bark' meanings of *bark*. Likewise, there is clustering in the t-SNE space of the different meanings of *shade*, although these different meanings also show greater overlap than in the case of *bark*. These differences in the geometry of multidimensional representations are consistent with previous verbal descriptions of the representational structure of homonymous and polysemous words (e.g. Rodd, 2020).

The multidimensional representations of *bark* and *shade* are visualised more precisely in Figure 5 using proxigrams. These overlay graphs use coloured lines to illustrate the nearest neighbours of each point and their distance in the high dimensional space (Kou, 2016). They mitigate potential artifacts of the t-SNE space such as false-neighbour and tear-up distortions by displaying more information that characterises the multidimensional space while maintaining the readability of a two-dimensional space (Bushati et al., 2011; Kou, 2016). Red lines connect points that are close to each other in the multidimensional space while blue lines highlight points that are further apart. The proxigram overlays of *bark* and *shade* highlight similar differences in their group structures. Whereas the proxigram of the contextualized embeddings of *shade* identifies multiple interconnected groups, the one for *bark* highlights two separate groups with very few connections between them, supporting previous verbal descriptions of these lexicographic categories (e.g. Rodd, 2020).

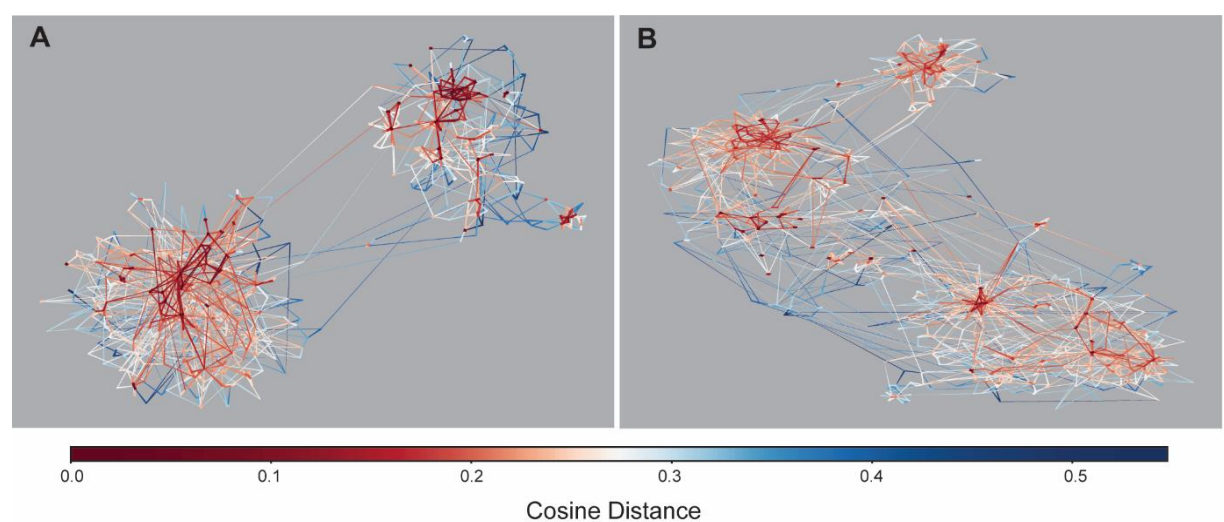

**Figure 5.** Proxigram overlays of the contextual embeddings of *bark* (panel A) and *shade* (panel B).

To evaluate quantitatively the degree to which the contextual embeddings for *bark* and *shade* contain meaningful, disambiguating information in the multidimensional space, we computed the intra- and inter-group cosine similarity of their contextual embeddings (with 'group' defined by translation labels; see Figure 4c). This analysis revealed greater similarity within each group than between groups ($b = -0.03, SE < 0.01, \beta = -0.31, t = -186.27, p < 0.001; \mu_{between} = 0.50; \mu_{within} = 0.56$). Moreover, we found a significant interaction between group status and type of ambiguity on the similarity of contextual embeddings ($b = -0.01, SE < 0.01, \beta = -0.08, t = -46.4, p < 0.001$). The similarity between groups is higher for *shade* than for *bark*, but the similarity within groups is higher for *bark* than *shade* ($\mu_{between-bark} = 0.49; \mu_{between-shade} = 0.51; \mu_{within-bark} = 0.57; \mu_{within-shade} = 0.55$). This analysis supports the visualisations in Figures 4 and 5, in revealing different geometries for homonymous and polysemous words. The learned contextual embeddings for homonymous *bark* show tightly bound, strongly separated meaning clusters, while contextual embeddings for polysemous *shade* show more diffuse interrelated, overlapping meaning clusters.

The analyses thus far demonstrate that the representations of *bark* and *shade* learned through the text input available to BERT show evidence of meaningful distinctions within the multidimensional space. Our final analysis used a Bayesian classifier to test the extent to which the contextual embeddings of *bark* and *shade* permit disambiguation of the multiple meanings of these words. We trained a Gaussian Naive Bayes algorithm (naïve_bayes.GaussianNB from the python sci-kit learn library; Pedregosa et al., 2011) on half of the contextual embeddings for *bark* and *shade* based on the translation labels (two for *bark*, 14 for *shade*). The trained classifier was then used to guess the meanings of the other half of the contextual embeddings for *bark* and *shade* that were not included in the training. The training and test subsets were randomly selected (using the train_test_split function with test_size=0.5 and random_state=0; sci-kit learn library; Pedregosa et al., 2011). The classifier had an accuracy of 94% and 74% in distinguishing the meanings of *bark* and *shade* respectively.

To avoid any potential for circularity (e.g. if the translation algorithm is based on technology similar to BERT), we repeated the classification analysis using ground truth labels derived from the human judgments described above. We took as ground truth the meanings selected for each context by at least two of our human raters (where this was not 'other') and excluded all other cases. This process left a total of 499 *bark* contexts and 465 *shade* contexts split between training and testing. The classification accuracy under these conditions was even higher than previously: 98% for *bark* and 84% for *shade*.

## Discussion

The results of our case study analyses indicate that BERT's representations of *bark* and *shade* in context possess a meaningful structure. The multidimensional representations of *bark* in context exhibit distinct, largely non-overlapping clusters of meaning, while multidimensional representations of *shade* in context exhibit multiple meaningful clusters characterised by significant overlap. The different geometries of these words align with lexicographic characterisations and verbal hypotheses regarding the nature of homonymous and polysemous words (e.g. Rodd, 2020). Our work is important because it provides computational evidence that these geometries (that are a cornerstone of psychological theories of lexical ambiguity; e.g. Rodd, 2020) emerge through the analysis of relatively superficial properties of text usage, and therefore may be learned straightforwardly by individuals through text experience.

Our classification analysis suggests that the meaningful information encoded in learned multidimensional representations of *bark* and *shade* has sufficient fidelity to support disambiguation at a high level of accuracy. This result is striking because BERT does not have deep, grounded knowledge of the meanings of these words. The information encoded in its representations is recovered only through BERT's experience of words and their relationship to other words in text. This result suggests that the written language contexts in which ambiguous words occur must provide extremely powerful disambiguating information. It may be that humans bring deep background knowledge and analysis to the problem of lexical ambiguity; however, our analyses suggest that this may not be required.

## General Discussion

Lexical ambiguity presents a profound and enduring challenge to the language sciences. For decades, researchers have grappled with understanding how language users select the right meaning when they encounter an ambiguous word (Johnson-Laird, 1987;

Swinney, 1979), and the cognitive challenges assumed to be associated with this process have led others to consider why languages admit ambiguity at all (Gibson et al., 2019; Piantadosi et al., 2012). Modern theories have made substantial progress in describing the processing dynamics involved in understanding ambiguous words (Rodd, 2020; Rodd et al., 2004). However, these theories are underpinned by an *assumption* about the representational structure of ambiguous words that has not been properly scrutinised. Specifically, these theories postulate that homonymous, polysemous, and unambiguous words have different geometries (Figure 1), but it is unclear how these different geometries might arise. Our work capitalises on advances in natural language models to gain new insight into this issue.

Our simulations interrogated the high-dimensional representations of ambiguous words in BERT (Devlin et al., 2019) a distributional semantic model that represents words as they arise in specific contexts. Our first simulation analysed the semantic spread of words previously characterised as homonymous, polysemous, or unambiguous (Armstrong & Plaut, 2016; Rodd et al., 2002), using BERT representations of these words in over 2.5 million contexts. Results showed that contextual representations of homonyms and polysemes show greater semantic spread than contextual representations of unambiguous words. That is, the similarity between each instance of an unambiguous word in context is greater than the similarity between each instance of an ambiguous word in context. This result suggests that ambiguous words may occur in more diverse contexts than unambiguous words, and therefore, that the representational structure depicted in Figure 1 may emerge through linguistic experience. However, this result by itself does not show that variation in semantic spread is meaningful (i.e. that distant representations have different meanings). Moreover, it provides only gross information about the geometry of representations; it says nothing about the extent to which homonymous and polysemous words are characterised by overlapping or non-overlapping representations (e.g. Rodd, 2020; see Figure 1).

Our second simulation therefore took a deep dive into the BERT representations of homonymous *bark* and polysemous *shade* using 500 randomly-selected contexts for each word. The visualisations presented in Figures 4 and 5 revealed different geometries for these words consistent with the schematic presented in Figure 1. These impressions were backed by quantitative analyses suggesting that homonymous *bark* is characterised by tightly-bound, non-overlapping clusters while polysemous *shade* is characterised by multiple overlapping clusters. Results from a Bayesian classification analysis suggested that information held within contextual embeddings of these words is sufficiently strong to disambiguate their multiple meanings, not only for homonymous *bark* (98% accuracy, 2 meanings) but also for polysemous *shade* (84% accuracy, 14 meanings) against human judgments. This performance is not perfect; however human raters also fail to agree on a proportion of these contexts. This work goes well beyond a previous demonstration that meaningful distinctions in contextual embeddings have higher fidelity than random meaning labels in words with two or three meanings (Li & Joanisse, 2021).

These results are novel and important because they suggest that the meaningful representations core to psychological understanding of ambiguity (Rodd, 2020) may be acquired through relatively low-level analysis of how these words are used. It is particularly striking that the linguistic distinction between homonymy and polysemy appears to be captured in the geometries of these representations, a finding that to our knowledge has not been demonstrated previously. These results immediately raise the question of how human language users acquire these representations. It may be that humans use deeper analysis of a broader range of inputs; however, our analyses suggest only a limited need to postulate more complex analytical processes. It will be for future work to determine the circumstances under which analysis of statistical patterns of usage fails to capture meaningful distinctions that are visible to human language users.

The approach to quantifying ambiguity developed in this work also offers opportunity to address a host of novel questions; for example, to study the geometry of representations in a wider range of ambiguous words (e.g. hybrids), to study the representations that emerge from analysis of different input corpora (e.g. children's literature), or to study the emergence of new meanings over time. The methods that we have introduced provide a means to quantify aspects of the different geometries of the semantic representations of words derived through distributional semantic models, and to move toward a characterisation of lexical ambiguity that is not based on strict lexicographic categories (see also Hoffman et al., 2013; Cevoli et al., 2021).

Our work provides initial evidence that the ambiguity structure of words may be recovered from relatively superficial analyses of usage. This finding implies that the text contexts in which we experience ambiguous words may be much more disambiguating than psycholinguists might have thought. The stimuli used in psycholinguistic studies of lexical ambiguity are often deliberately structured to make disambiguation challenging: for example, consider the following sentence stimuli from Duffy et al. (1988).

> "*Of course, the boxer was exhausted by the time they got it back on its leash.*"

> "*Last night the cabinet was covered with dust after spending the day talking with miners in the mines.*"

We need to remember that these types of sentences may not reflect the natural language contexts in which ambiguous words occur. Indeed, the fact that BERT's learned representations hold strongly disambiguating information implies that the contexts in which these words occur are themselves strongly disambiguating. Lexical ambiguity may not be the substantial cognitive challenge that psychological theorising has assumed it to be.

The novel insights that have emerged through our simulations have only become possible because of recent advances in natural language processing, and specifically because of models that represent unique instances of words differently, in terms of their contexts. In effect, these models realise Firth's insight from almost one hundred years ago: "*...the complete meaning of a word is always contextual, and no study of meaning apart from a complete context can be taken seriously*" (Firth, 1935). This characterisation is a radical departure from the notion that the meanings of words are "*highly stable states*" (Rodd, 2020, p. 414). It will therefore be important to evaluate whether a radical contextual theory of meaning like BERT (Devlin et al., 2019) provides a viable account of human language understanding.

Such a discussion is beyond the scope of this article; however, we would remind readers of a playful study demonstrating that just a few sentences of context can overcome the primary meanings of well-known words (Nieuwland & Van Berkum, 2006). Participants heard stories featuring an inanimate object like that printed below while EEG data were recorded.

> "*A woman saw a dancing peanut who had a big smile on his face. The peanut was singing about a girl he had just met. And judging from the song, the peanut was totally crazy about her. The woman thought it* *was really cute to see the peanut singing and dancing like that. The peanut was **salted / in love …***"

The critical finding was that the N400 was larger (indicating processing difficulty) given the target '*salted*' versus '*in love*'. Just four sentences of context turned the lifelong association between *peanut* and *salted* into a semantic anomaly. If the meaning of peanut can change so rapidly, based on so little information, then how might that be captured as part of a "stable state"? Remarkably, when BERT 'reads' the passage above up to the final sentence "*The peanut was …*", its top-5 predictions for the next word are *singing*, *talking*, *going*, *just*, and *dancing*. Likewise, when each instance of the word *peanut* is masked one at a time, BERT's predictions indicate that the model interprets the peanut as a human (i.e. *person*, *man*, *guy*).

To conclude, we have capitalised on advances in natural language modelling to gain new insights into human understanding of ambiguous words. We have used a model that has no grounded understanding of the meanings of words at all; it simply predicts words based on the surrounding context provided by other words. Remarkably, it's contextual representations appear to cluster according to the senses in which words are used, with a degree of precision that supports highly-accurate classification along the finest shades of meaning visible to a lexicographer. It seems unlikely that human language understanding reflects only the superficial information available to the model; however, the modelling does provide a baseline for probing how the deep semantic knowledge characteristic of human language processing adds value to this lower-level statistical information.

## References

Aiken, M. (2019). An updated evaluation of google translate accuracy. *Studies in Linguistics and Literature*, *3*(3), 253. https://doi.org/10.22158/SLL.V3N3P253

Allen, R. (1999). Lumping and splitting. *English Today*, *15*(4), 61–63. https://doi.org/10.1017/S0266078400011317

Armstrong, B. C., & Plaut, D. C. (2008). Settling dynamics in distributed networks explain task differences in semantic ambiguity effects: Computational and behavioral evidence.

*Proceedings of the Annual Meeting of the Cognitive Science Society*, *30*(30). https://escholarship.org/content/qt6rv2c5hh/qt6rv2c5hh.pdf

Armstrong, B. C., & Plaut, D. C. (2016). Disparate semantic ambiguity effects from semantic processing dynamics rather than qualitative task differences. *Language, Cognition and Neuroscience*, *31*(7), 940–966. https://doi.org/10.1080/23273798.2016.1171366

Arora, S., Li, Y., Liang, Y., Ma, T., & Risteski, A. (2018). Linear algebraic structure of word senses, with applications to polysemy. *Trans. Assoc. Comput. Linguist.*, *6*, 483–495. https://doi.org/10.1162/tacl_a_00034

Baccouri, N. (2020). *Deep Translator - PyPI*. https://pypi.org/project/deep-translator/

Bartsch, R. (1984). Norms, tolerance, lexical change, and context- dependence of meaning. *Journal of Pragmatics*, *8*(3), 367–393. https://doi.org/10.1016/0378-2166(84)90029-8

Beekhuizen, B., Armstrong, B. C., & Stevenson, S. (2021). Probing lexical ambiguity: word vectors encode number and relatedness of senses. *Cognitive Science*, *45*(5), e12943. https://doi.org/10.1111/cogs.12943

Bender, E. M., Gebru, T., McMillan-Major, A., & Shmitchell, S.-G. (2021). On the dangers of stochastic parrots: can language models be too big? *FAccT 2021 - Proceedings of the 2021 ACM Conference on Fairness, Accountability, and Transparency*, 610–623. https://doi.org/10.1145/3442188.3445922

Burgess, C. (1998). From simple associations to the building blocks of language: Modeling meaning in memory with the HAL model. *Behavior Research Methods, Instruments, and Computers*, *30*(2), 188–198. https://doi.org/10.3758/BF03200643

Bushati, N., Smith, J., Briscoe, J., & Watkins, C. (2011). An intuitive graphical visualization technique for the interrogation of transcriptome data. *Nucleic Acids Research*, *39*(17), 7380–7389. https://doi.org/10.1093/nar/gkr462

Camacho-Collados, J., & Pilehvar, M. T. (2018). From word to sense embeddings: A survey on vector representations of meaning. *Journal of Artificial Intelligence Research*, *63*, 743–788. https://doi.org/10.1613/jair.1.11259

Cevoli, B., Watkins, C., & Rastle, K. (2021). What is semantic diversity and why does it facilitate visual word recognition? *Behavior Research Methods*, *53*(1), 247–263. https://doi.org/10.3758/s13428-020-01440-1

Cevoli, B., Watkins, C., & Rastle, K. (2022). Prediction as a basis for skilled reading: insights from modern language models. *Royal Society Open Science*, *9*(6). https://doi.org/10.1098/RSOS.211837

Devlin, J., Chang, M. W., Lee, K., & Toutanova, K. (2019). BERT: Pre-training of deep bidirectional transformers for language understanding. *NAACL HLT 2019 - 2019 Conference of the North American Chapter of the Association for Computational Linguistics: Human Language Technologies - Proceedings of the Conference*, *1*, 4171–4186.

Duffy, S. A., Morris, R. K., & Rayner, K. (1988). Lexical ambiguity and fixation times in reading. *Journal of Memory and Language*, *27*(4), 429–446. https://doi.org/10.1016/0749-596X(88)90066-6

Fang, X., Perfetti, C., & Stafura, J. (2017). Learning new meanings for known words: Biphasic effects of prior knowledge. *Language, Cognition and Neuroscience*, *32*(5), 637–649. https://doi.org/10.1080/23273798.2016.1252050

Firth, J. R. (1935). The technique of semantics. *Transactions of the Philological Society*, *34*(1), 36–73. https://doi.org/10.1111/J.1467-968X.1935.TB01254.X

Gibson, E., Futrell, R., Piandadosi, S. T., Dautriche, I., Mahowald, K., Bergen, L., & Levy, R. (2019). How efficiency shapes human language. *Trends in Cognitive Sciences*, *23*(5), 389–407. https://doi.org/10.1016/j.tics.2019.02.003

Gilliver, P. (2016). *The making of the Oxford English dictionary* (Illustrated). OUP Oxford.

Google. (2021). *Google Translate*. https://translate.google.com/?hl=en&sl=en&tl=it&text=shade&op=translate

Hino, Y., Lupker, S. J., & Pexman, P. M. (2002). Ambiguity and synonymy effects in lexical decision, naming, and semantic categorization tasks: Interactions between orthography, phonology, and semantics. *Journal of Experimental Psychology: Learning Memory and Cognition*, *28*(4), 686–713. https://doi.org/10.1037/0278-7393.28.4.686

Hoffman, P., Lambon Ralph, M. A., & Rogers, T. T. (2013). Semantic diversity: A measure of semantic ambiguity based on variability in the contextual usage of words. *Behavior Research Methods*, *45*(3), 718–730. https://doi.org/10.3758/s13428-012-0278-x

Hoffman, P., Lambon Ralph, M. A., & Rogers, T. T. (2021). Semantic diversity is best measured with unscaled vectors: Reply to Cevoli, Watkins and Rastle (2020). *Behavior Research Methods*, 1–13. https://doi.org/10.3758/S13428-021-01693-4/TABLES/6

Hsiao, Y., & Nation, K. (2018). Semantic diversity, frequency and the development of lexical quality in children's word reading. *Journal of Memory and Language*, *103*, 114–126. https://doi.org/10.1016/J.JML.2018.08.005

Johnson-Laird, P. N. (1987). The mental representation of the meaning of words. *Cognition*, *25*(1–2), 189–211. https://doi.org/10.1016/0010-0277(87)90009-6

Kawamoto, A. H., Farrar, W. T., & Kello, C. T. (1994). When two meanings are better than one: Modeling the ambiguity advantage using a recurrent distributed network. *Journal of Experimental Psychology: Human Perception and Performance*, *20*(6), 1233–1247. https://doi.org/10.1037/0096-1523.20.6.1233

Kilgarriff, A. (1997). I don't believe in word senses. *Computers and the Humanities 1997 31:2*, *31*(2), 91–113. https://doi.org/10.1023/A:1000583911091

Klein, D. E., & Murphy, G. L. (2001). The representation of polysemous words. *Journal of Memory and Language*, *45*, 259–282. https://doi.org/10.1006/jmla.2001.2779

Klepousniotou, E., Titone, D., & Romero, C. (2008). Making sense of word senses: The comprehension of polysemy depends on sense overlap. *Journal of Experimental Psychology: Learning Memory and Cognition*, *34*(6), 1534–1543. https://doi.org/10.1037/a0013012

Kou, J. (2016). *Faithful visualisation of similarities in high dimensional data*.

Kwon, S., Kang, C., Han, J., & Choi, J. (2019). Why do masked neural language models still need common sense knowledge? *ArXiv*.

Landauer, T. K., & Dumais, S. T. (1997). A solution to Plato's problem: the latent semantic analysis theory of acquisition, induction, and representation of knowledge. *Psychological Review*, *104*(2), 211–240. https://doi.org/10.1037/0033-295X.104.2.211

Li, J., & Joanisse, M. F. (2021). Word senses as clusters of meaning modulations: a computational model of polysemy. *Cognitive Science*, *45*(4), e12955. https://doi.org/10.1111/COGS.12955

Marcus, G., & Davis, E. (2019). *Rebooting AI: building artificial intelligence we can trust*. Knopf Doubleday Publishing Group.

Martin, D. I., & Berry, M. W. (2013). Mathematical foundations behind Latent Semantic Analysis. In T. K. Landauer, D. S. McNamara, S. Dennis, & W. Kintsch (Eds.), *Handbook of Lantent Semantic Analysis* (pp. 35–56). Routledge.

McClelland, J. L., Hill, F., Rudolph, M., Baldridge, J., & Schütze, H. (2019). Extending machine language models toward human-level language understanding. *ArXiv*.

McDonald, S. A., & Shillcock, R. C. (2001). Rethinking the word frequency effect: The neglected role of distributional information in lexical processing. *Language and Speech*, *44*(3), 295–322. https://doi.org/10.1177/00238309010440030101

Merity, S., Keskar, N. S., & Socher, R. (2017). Regularizing and optimizing LSTM language models. *6th International Conference on Learning Representations, ICLR 2018 - Conference Track Proceedings*.

Mikolov, T., Chen, K., Corrado, G., & Dean, J. (2013). Efficient estimation of word representations in vector space. *ArXiv*. http://arxiv.org/abs/1301.3781

Mu, J., Bhat, S., & Viswanath, P. (2017). Geometry of polysemy. In G. Balint, B. Antala, C. Carty, J.-M. A. Mabieme, I. B. Amar, & A. Kaplanova (Eds.), *5th International Conference on Learning Representations, ICLR 2017*. https://doi.org/10.2/JQUERY.MIN.JS

Nieuwland, M. S., & Van Berkum, J. J. A. (2006). When peanuts fall in love: N400 evidence for the power of discourse. *Journal of Cognitive Neuroscience*, *18*(7), 1098–1111. https://doi.org/10.1162/jocn.2006.18.7.1098

Pedregosa, F., Varoquaux, G., Gramfort, A., Michel, V., Thirion, B., Grisel, O., Blondel, M., Prettenhofer, P., Weiss, R., Dubourg, V., Vanderplas, J., Passos, A., Cournapeau, D., Brucher, M., Perrot, M., & Duchesnay, É. (2011). Scikit-learn: machine learning in python. *Journal of Machine Learning Research*, *12*(85), 2825–2830.

Pexman, P. M., Hino, Y., & Lupker, S. J. (2004). Semantic ambiguity and the process of generating meaning from print. *Journal of Experimental Psychology: Learning Memory and Cognition*, *30*(6), 1252–1270. https://doi.org/10.1037/0278-7393.30.6.1252

Piantadosi, S. T., Tily, H., & Gibson, E. (2012). The communicative function of ambiguity in language. *Cognition*, *122*(3), 280–291. https://doi.org/10.1016/j.cognition.2011.10.004

Rodd, J. M. (2020). Settling into semantic space: an ambiguity-focused account of word-meaning access. *Perspectives on Psychological Science*, *15*(2), 411–427. https://doi.org/10.1177/1745691619885860

Rodd, J. M., Gaskell, G., & Marslen-Wilson, W. (2002). Making sense of semantic ambiguity: Semantic competition in lexical access. *Journal of Memory and Language*, *46*(2), 245–266. https://doi.org/10.1006/jmla.2001.2810

Rodd, J. M., Gaskell, M. G., & Marslen-Wilson, W. D. (2004). Modelling the effects of semantic ambiguity in word recognition. *Cognitive Science*, *28*(1), 89–104. https://doi.org/10.1016/j.cogsci.2003.08.002

Swinney, D. A. (1979). Lexical access during sentence comprehension: (Re)consideration of context effects. *Journal of Verbal Learning and Verbal Behavior*, *18*(6), 645–659. https://doi.org/10.1016/S0022-5371(79)90355-4

Tabossi, P. (1988). Accessing lexical ambiguity in different types of sentential contexts. *Journal of Memory and Language*, *27*(3), 324–340. https://doi.org/10.1016/0749-596X(88)90058-7

The British National Corpus. (2007). *Version 3 (BNC XML Edition)*. Distributed by Bodleian Libraries, University of Oxford, on behalf of the BNC Consortium.

Tuggy, D. (1993). Ambiguity, polysemy, and vagueness. *Cognitive Linguistics*, *4*(3), 273–290. https://doi.org/10.1515/COGL.1993.4.3.273/MACHINEREADABLECITATION/RIS

Van Der Maaten, L., & Hinton, G. (2008). Visualizing data using t-SNE. In *Journal of Machine Learning Research* (Vol. 9).

Wittgenstein, L. (1958). *Preliminary studies for the "Philosophical Investigations", generally known as the blue and brown books*. Blackwell Publishers Ltd.

Wolf, T., Debut, L., Sanh, V., Chaumond, J., Delangue, C., Moi, A., Cistac, P., Rault, T., Louf, R., Funtowicz, M., Davison, J., Shleifer, S., von Platen, P., Ma, C., Jernite, Y., Plu, J., Xu, C., Scao, T. le, Gugger, S., … Rush, A. M. (2019). *HuggingFace's Transformers: State-of-the-art Natural Language Processing*. https://doi.org/10.48550/arxiv.1910.03771